# MedicalBERT: enhancing biomedical natural language processing using pretrained BERT-based model

K. Sahit Reddy, N. Ragavenderan, Vasanth K., Ganesh N. Naik, Vishalakshi Prabhu, Nagaraja G. S.
Department of Computer Science, R. V. College of Engineering, Bengaluru, India

**ABSTRACT**

Recent advances in natural language processing (NLP) have been driven by pretrained language models like BERT, RoBERTa, T5, and GPT. These models excel at understanding complex texts, but biomedical literature, with its domain-specific terminology, poses challenges that models like Word2Vec and bidirectional long short-term memory (Bi-LSTM) can't fully address. GPT and T5, despite capturing context, fall short in tasks needing bidirectional understanding, unlike BERT. Addressing this, we proposed MedicalBERT, a pretrained BERT model trained on a large biomedical dataset and equipped with domain-specific vocabulary that enhances the comprehension of biomedical terminology. MedicalBERT model is further optimized and fine-tuned to address diverse tasks, including named entity recognition, relation extraction, question answering, sentence similarity, and document classification. Performance metrics such as the F1-score, accuracy, and Pearson correlation are employed to showcase the efficiency of our model in comparison to other BERT-based models such as BioBERT, SciBERT, and ClinicalBERT. MedicalBERT outperforms these models on most of the benchmarks, and surpasses the general-purpose BERT model by 5.67% on average across all the tasks evaluated respectively. This work also underscores the potential of leveraging pretrained BERT models for medical NLP tasks, demonstrating the effectiveness of transfer learning techniques in capturing domain-specific information.

**Corresponding Author:**

N. Ragavenderan
Department of Computer Science, R. V. College of Engineering
Mysore Rd, RV Vidyaniketan Post, Bengaluru, India
Email: nragavenderan.cs22@rvce.edu.in

## 1. INTRODUCTION

In the realm of natural language processing (NLP), the revolutionary influence of sophisticated transformer models is observable across multiple fields. In the healthcare industry, electronic health records (EHRs) are a substantial source of valuable data, and BERT has proven to be a powerful tool for deciphering the intricate contextual details embedded within unstructured medical text.

BERT [1], short for bidirectional encoder representations from transformers, is a formidable NLP model that was introduced by Google in 2018. Its key innovation lies in its bidirectional approach, which enables it to take into account the context of words from both directions. Pretrained on extensive text data, BERT excels in capturing nuanced language relationships, making it a cornerstone in various NLP applications. In the legal domain, BERT assists in parsing and extracting relevant information from legal documents. While BERT has demonstrated proficiency in general language understanding, the healthcare domain demands a specialized approach.

Conventional NLP techniques, like bag-of-words (BoW) [2] and n-grams [3], struggle with contextual understanding, treating words independently and often leading to information loss, especially with varying text lengths. Methods like TF-IDF also face issues with polysemy, manual feature engineering, and capturing long-range dependencies, limiting performance and transfer learning. Transformer models like BERT solve these issues by using deep bidirectional contextual representations, improving adaptability and accuracy across diverse NLP tasks.

This work focuses on proposing a BERT-based model, MedicalBERT, for biomedical text mining and NLP, with special attention given to the unique characteristics that differentiate this model from other specialized BERT models in this domain. Through experimentation and evaluation by pre-training and fine-tuning, we demonstrate how this proposed model stands out, by comparing the outcome of our model, with other BERT-based models that once emerged as state-of-the-art models and contributed to the BioNLP community. Section 5 presents a comparative analysis and results of our proposed model with those of other models. Pretraining refers to the initial phase where the model learns language patterns from a large corpus of text data, capturing general language features. Fine-tuning, on the other hand, involves further training the pretrained model on a specific task or domain to enhance its performance for that particular application. Detailed evaluation metrics illustrate the model's strengths and highlight areas for potential improvement. This benchmarking not only demonstrates our proposed model's improvements but also provides insights into the specific strengths and weaknesses relative to existing models. This work also highlights the significance of using custom tailored vocabulary and a larger training set and its impact on the performance of the model.

This paper consists of 6 sections–1$^{st}$ section is the introduction, followed by the 2$^{nd}$ section which gives a brief knowledge about the related BERT-based models that will be used for performance comparison presented in the results section, i.e., 5$^{th}$ section. The 3$^{rd}$ section gives an overview of BERT architecture, and all the benchmark datasets and the corresponding NLP tasks performed on each. Continuing forward, the 4$^{th}$ section provides the method in which MedicalBERT was trained and fine-tuned to give the final proposed model, which outperforms most of the state-of-the-art models in the biomedical domain. Finally, the 5$^{th}$ section provides the results of our work and further discussion about the work, followed by the conclusion section.

## 2. RELATED MODELS AND WORKS
### 2.1. BioBERT

BioBERT [4], a domain-specific linguistic portrayal framework, emerges from an initial BERT checkpoint trained on biomedical text from PubMed Abstracts and PubMed Central. Its adaptation for biomedical text mining acknowledges the growing importance of this field, driven by the rapid expansion of biomedical document volumes. NLP advancements applied directly to biomedical text mining often yield suboptimal outcomes due to distribution shifts from general domain corpora to biomedical corpora. BioBERT outperforms BERT and earlier top-performing models across various biomedical text mining tasks, including named entity recognition, relation extraction, and question answering. Although effective in general biomedical NLP tasks, BioBERT's reliance on a more generalized vocabulary occasionally reduces its performance on datasets with highly specific terminology. In contrast, MedicalBERT uses a custom biomedical vocabulary aligned with BERT tokenization, allowing it to better understand and interpret complex terms unique to biomedical literature.

### 2.2. BlueBERT

Inspired by the success of the general language understanding evaluation benchmark, this work introduced the biomedical language understanding evaluation (BLUE) benchmark [5]. Evaluations on multiple baselines involving BERT [1] and ELMo [6] provide a comparative analysis of pretraining language representations in the biomedical field. The BERT model pretrained on PubMed abstracts and MIMIC-III clinical notes achieved the best results among the evaluated baselines, extending benchmarking success to biomedicine. The BLUE benchmark comprises five assignments with ten datasets, encompassing a variety of biomedical and clinical documents, including various dataset sizes and difficulties. By focusing narrowly on PubMed and MIMIC-III data, BlueBERT may capture patterns and terminology specific to these datasets, potentially limiting its generalizability across broader biomedical or clinical texts. This can lead to reduced performance on datasets that use diverse or less-common medical vocabulary, as the model may have learned patterns that are overly specific to its pretraining data. In comparison, MedicalBERT's pretraining on extensive biomedical data addresses these issues, showing robust performance across varied biomedical benchmarks without the same overfitting limitations.

### 2.3. ClinicalBERT

ClinicalBERT [7], pretrained on clinical notes, uncovers high-quality relationships between medical concepts, maximizing the use of clinical documentation. It develops and evaluates a continuous representation of high-dimensional and sparse clinical notes, estimating the likelihood of patients being readmitted to the hospital within 30 days at different intervals, leveraging the BERT model. Surpassing multiple baselines in forecasting hospital readmissions within a 30-day timeframe, utilizing discharge summaries and preliminary notes from the intensive care unit, ClinicalBERT's effectiveness in representing clinical notes may vary across different clinical domains, requiring further validation. However, its performance varies across clinical domains due to the diversity in medical language and terminology used in different specialties. For instance, while it excels with general patient data, tasks involving niche medical fields or rare clinical conditions may show reduced accuracy. More extensive validation on specialized datasets would help assess ClinicalBERT's adaptability across various clinical sub-domains, highlighting where MedicalBERT's broad biomedical pretraining offers a more robust alternative for diverse clinical and biomedical applications.

### 2.4. SciBERT

Given the scarcity of extensively annotated data for NLP tasks in the scientific domain, SciBERT [8], a pretrained language model built upon BERT, is introduced. Leveraging unsupervised pretraining on a diverse, multidisciplinary collection of scientific publications and citations enhances its effectiveness on downstream scientific NLP tasks. Employing its own vocabulary called SciVocab, consisting of scientific terms, SciBERT shows significant statistical enhancements compared to BERT on diverse scientific NLP assignments, encompassing sequence tagging, sentence categorization, and dependency analysis. While SciBERT effectively handles general scientific literature, its performance in the biomedical domain is limited due to vocabulary constraints and lack of specific biomedical training data. MedicalBERT, on the other hand, is pretrained specifically on biomedical corpus. This specialization allows MedicalBERT to achieve higher accuracy in tasks such as named entity recognition and relation extraction, particularly on datasets like BC5CDR-Chemical and NCBI-Disease, where it outperforms SciBERT by an F1 margin of 1.37 on average. MedicalBERT's domain-specific vocabulary enhances its understanding of complex biomedical terms, addressing a notable gap in SciBERT's design for such applications.

### 2.5. RoBERTa

Robustly optimized BERT (RoBERTa) [9] is a BERT model that underwent training on an expanded English dataset and for a longer period of time using self-supervised training techniques. This results in a better model for various NLP tasks. RoBERTa's extensive pretraining on general English text allows it to perform reasonably well on broad biomedical tasks; however, it falls short in deeper domain-specific contexts. For instance, while RoBERTa demonstrates competitive performance on the LINNAEUS dataset with an F1 score of 87.8, it underperforms on more complex biomedical tasks. MedicalBERT, in contrast, is fine-tuned specifically for biomedical text mining, outperforming RoBERTa in tasks requiring higher contextual and domain-specific comprehension, such as in the BC5CDR-Disease dataset, where MedicalBERT shows a 2.01 F1 score improvement over RoBERTa. This improvement underscores MedicalBERT's advantage in leveraging biomedical terminology and transfer learning for specialized tasks.

### 2.6. Other works

Bressem *et al*. [10] introduced, a pre-trained BERT model using a large corpus of German medical documents, including radiology reports, PubMed abstracts, Springer Nature, and German Wikipedia. Preprocessing steps involved data anonymization (removing patient names with a named entity recognition model) and deduplication (using cosine similarity). The model achieved state-of-the-art performance on eight medical benchmarks, largely due to the extensive training data, with efficient tokenization having a lesser effect on results.

Wada *et al*. [11] tackled BERT's weaker performance on smaller biomedical corpora by up-sampling smaller datasets and pre-training on both large and small corpora. This involved segmenting both corpora into similar-sized documents and increasing the smaller corpus to match the larger one. Three experiments were conducted: a Japanese medical BERT, an English biomedical BERT, and an enhanced biomedical BERT from PubMed. Results showed the Japanese BERT excelled in medical classification tasks, while the English BERT performed well on the BLUE benchmark, with the enhanced BERT model improving scores by 0.3 points over the ablation model.

## 3. OVERVIEW OF BERT AND BENCHMARKS
### 3.1. BERT

BERT are a transformer-derived language representation model designed to understand and capture the contextual significance of words within a provided sentence or passage by capturing bidirectional relationships between words [1]. BERT employs a transformer architecture based on self-attention mechanisms [12]. Contrary to recurrent neural networks (RNNs), long short-term memory (LSTM), and other models based on unidirectional attention that scrutinize text either from left to right or from right to left [13], [14], the architecture of BERT enables it to understand the relationships between words within a sentence by processing text in both directions, both leftward and rightward, by processing the entire sequence simultaneously. By employing two unsupervised learning tasks, namely masked language modeling (MLM) and next sentence prediction (NSP), BERT can engage in pretraining on extensive amounts of unlabeled text tailored to a specific domain [1].

In MLM, each token is converted into word embeddings. In the input sequence, a certain percentage (~15%) of the input tokens are randomly chosen and obscured to be substituted with a special [MASK] token (~80%). The model handles the input sequence token by token using self-attention mechanisms and multilayer neural networks [13]. For each masked token, the model generates a probability distribution across the entire token vocabulary. By training the model to reconstruct the masked tokens, the model learns to produce coherent representations of the text [1]. NSP determines whether two sentences appear consecutively in a text corpus. The BERT model tags each instance to denote whether the subsequent sentence logically succeeds the preceding sentence in the source text. It tokenizes the sentences and creates embeddings for each token. Special tokens such as [CLS] (classification) and [SEP] (separator) are inserted between sentences, and segment embeddings are introduced to differentiate between the two sentences [1].

### 3.2. Downstream natural language processing tasks
#### 3.2.1. Named entity recognition

Conducting named entity recognition on biomedical datasets involves identifying and classifying entities like genes, diseases, proteins, and chemicals present in the text. Older models in this domain, such as conditional random fields (CRFs) [15], [16] and bidirectional long short-term memory (Bi-LSTM) [17], [18] networks, depend on manually crafted features or sequential dependencies but lack the ability to capture intricate context representations effectively.

BERT outperforms these older models by leveraging bidirectional context understanding and transformer architectures. The model learns to label each input token with its corresponding entity label or entity type using BIO tags (beginning, inside, outside), which our model employs for this task. The F1 metric and word-level macro F1 (applied to PICO task [19]) were utilized for result evaluation.

#### 3.2.2. Relation extraction

In the biomedical domain, relation extraction involves identifying and categorizing connections among entities mentioned in textual content, such as proteins, genes, diseases, and drugs. Methods like support vector machines (SVMs) and CRFs [15], [16], or convolutional neural networks (CNNs) endeavor to perform relation extraction by leveraging linguistic features, syntactic parsing, or sequence modeling. However, they struggled to capture complex semantic relationships effectively.

Annotated datasets with entities @GENE$ and @DISEASE$ in the genetic association database (GAD) and European Union adverse drug reactions (EU-ADR) datasets, @GENE and @CHEMICAL$ in the ChemProt dataset and @DRUG$ in the DDI dataset are employed for training purposes. Each sequence is formatted with special tokens, allowing BERT to understand the relationships between entities. These unique tokens denote the start and end of entities or relationships, namely [CLS] and [SEP]. The F1 and micro average-F1 metrics are employed to assess the outcomes of this task.

#### 3.2.3. Question answering

Question answering within the biomedical domain encompasses providing answers to queries formulated in natural language, relying on information contained within biomedical texts like research articles and clinical records. Information retrieval systems and the Java library Lucene utilized techniques such as term frequency-inverse document frequency (TF-IDF) and straightforward token matching for conducting question answering, but they were limited in capturing contextual and semantic understanding.

BERT learns by processing both the questions and context passages through its transformer-based architecture, involving several layers of self-attention and encoding mechanisms. We used two datasets from biomedical language understanding and reasoning benchmark (BLURB) [20]: PubMed question answering (PubMedQA) [21] and biomedical semantic indexing and question-answering challenge (BioASQ) [22] to perform the question-answering task and evaluated the results based on their accuracy.

#### 3.2.4. Sentence similarity

Sentence similarity involves encoding each sentence separately with BERT, extracting the embeddings corresponding to each token, and aggregating these embeddings into a fixed-size vector representation encompassing the entire sentence. After the sentence embeddings are acquired, cosine similarity is employed to gauge the similarity between the two vectors. A higher score for cosine similarity indicates greater similarity between sentences, while a lower score implies less similarity. We employed the biomedical semantic sentence similarity estimation system (BIOSSES) [23] dataset from BLURB and evaluated the performance on the Pearson metric.

### 3.2.5. Document multilabel classification

Document classification is an NLP task that involves assigning categories or classes to a document. This method simplifies the management, searching, filtering, and analysis of documents. We evaluated the benchmark on the hallmarks of cancer (HoC) dataset [24], which is also available in BLURB [20].

### 3.3. Datasets and benchmarks

We utilized existing datasets used for tasks in the BLURB [20] benchmark and BioBERT [4], which are commonly used by the BioNLP community. A total of 17 datasets were used, and Table 1 lists the training and testing instances for each of the five tasks, which include sequence labeling, sequence classification, question answering, semantic sentence similarity, and document classification.

Table 1. Statistics of the datasets utilized for fine-tuning. The training and test columns indicate the number of training and testing examples for each dataset respectively

| Dataset | Task | Train | Test |
| --- | --- | --- | --- |
| NCBI-disease | Named entity recognition | 5134 | 960 |
| BC2GM | Named entity recognition | 15197 | 6325 |
| BC5CDR-disease | Named entity recognition | 4182 | 4424 |
| BC5CDR-chemical | Named entity recognition | 5203 | 5385 |
| JNLPBA | Named entity recognition | 46750 | 8662 |
| LINNAEUS | Named entity recognition | 281273 | 165095 |
| BC4CHEMD | Named entity recognition | 893685 | 767636 |
| Species-800 | Named entity recognition | 147291 | 42298 |
| EBM-PICO | PICO | 339167 | 16364 |
| GAD | Relation extraction | 4261 | 535 |
| EU-ADR | Relation extraction | 3195 | 355 |
| DDI-2013 | Relation extraction | 22233 | 5716 |
| ChemProt | Relation extraction | 18035 | 15745 |
| PubMedQA | Question answering | 450 | 500 |
| BioASQ | Question answering | 670 | 140 |
| BIOSSES | Sentence similarity | 64 | 20 |
| HoC | Multilabel classification | 1295 | 371 |

NCBI-disease [25] is entirely annotated and contains 793 abstracts from PubMed, featuring 6,892 mentions of diseases and 790 distinct disease concepts. The BioCreative V CDR (BC5CDR) [26] dataset comprises lengthy documents that are segmented into sentences to mitigate their size, as they cannot be directly processed by language models due to size constraints. The two entity types are chemical and disease. The BioCreative IV chemical compound and drug name recognition (BC4CHEMD) [19] database contains 10,000 PubMed abstracts that contain 84,355 chemical entity mentions in total. They are labeled manually by chemistry literature curators.

The BioCreative 2 gene mention (BC2GM) [27] corpus is an aggregation of various sentences, where each sentence is composed of gene mentions (GENE annotations). Participants are tasked with pinpointing a gene mention within a sentence by identifying the starting and ending characters of that sentence. JNLPBA [28] is derived from the GENIA version 3.02 corpus. This tool was constructed based on a controlled search of MEDLINE using MeSH terms such as "blood cells" and "transcription factors". Approximately 2,000 abstracts were filtered and annotated by hand in accordance with a small taxonomy of 48 classes based on a chemical classification.

Species-800 [29] is a corpus whose abstracts are manually annotated. It is a corpus for specific entities. It consists of 800 PubMed abstracts that contain organism mentions. From eight disciplines, namely bacteriology, botany, entomology, medicine, mycology, protistology, virology, and zoology, 100 abstracts were chosen. LINNAEUS [30] corpus includes 100 full-text documents sourced randomly from the PMCOA document set. Species within these documents were annotated by hand and then standardized to the NCBI taxonomy IDs corresponding to the specific species mentioned.

EBM-PICO [31] used for the PICO task contains 4,993 abstracts. These are annotated with (P)articipants, (I)ntervention, (C)omparator, and (O)utcomes. GAD [32] is a corpus that is used to identify associations between genes and diseases. It uses a semiautomatic annotation procedure. EU-ADR [33] dataset has been annotated for disorders, drugs, their interrelationships and genes. To understand these relationships in texts, the annotated relationships serve as a basis for training and assessing text mining techniques. ChemProt [34] dataset identifies chemical and protein entities and their likely relation to one another. These compounds are generally activators or inhibitors of proteins.

PubMedQA [21] is a dataset based on biomedical question answering collected from PubMed abstracts. The objective of the model developed with this dataset provides answers to biomedical questions such as YES, NO, or MAYBE, based on information from the abstracts. The dataset has 1000 expert-annotated, 61,200 unlabeled and 211,300 artificially generated question answer instances. Each instance consists of a question that is based on the existing research title.

BioASQ [22] question answering dataset consists of, in addition to exact answers, ideal answers that are useful for research on multidocument summarization, unlike the majority of the previous question answering benchmarks, which consist of only exact answers. It consists of structured and unstructured data. It consists of documents and snippets that prove useful for retrieval of information and passage retrieval experiments. BIOSSES [23] is a benchmark dataset for biomedical sentence similarity estimation. The dataset contains 100 sentence pairs, selected from the text analysis conference (TAC) biomedical summarization track training dataset, which is composed of articles from the biomedical field. The sentence pairs included in the dataset were chosen from sentences that cite a reference article.

## 4. METHOD
### 4.1. Pretraining the BERT model

The original BERT (large) weights trained on the general domain, namely BooksCorpus and Wikipedia, were employed as a starting point [1]. To adapt BERT for biomedical text, pretraining was performed on a substantial volume of PubMed and MIMIC-III and clinical notes. The pre-training and fine-tuning processes are briefly depicted in Figure 1.

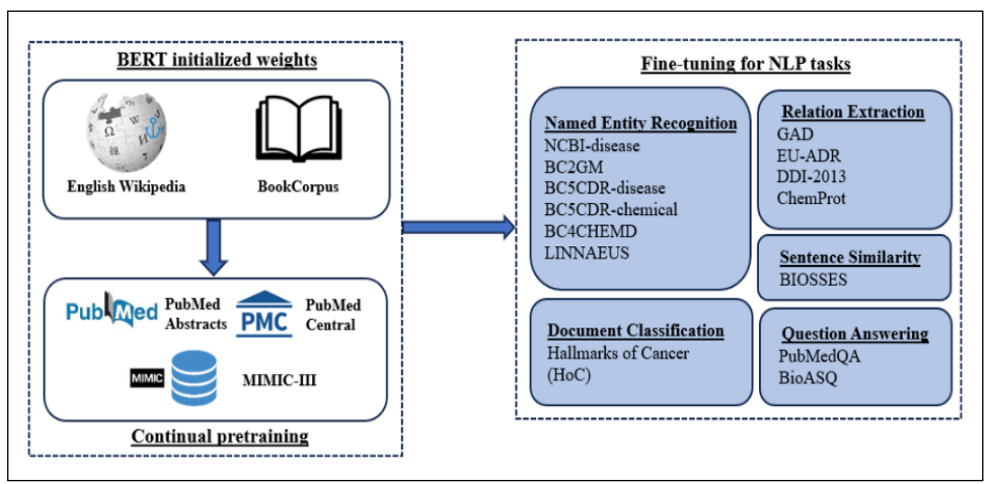

Figure 1. Overview of the pretraining and fine-tuning stages of MedicalBERT

The statistics of the pretraining corpora are given in Table 2. The WordPiece tokenization technique [34] which was employed by the original BERT, is a subword tokenization method, is well suited for biomedical corpora because it can effectively manage various types of vocabulary, including specialized medical terminology, acronyms, and rare or out-of-vocabulary words commonly found in biomedical texts. BioBERT [4] for tokenization. But it was observed that using domain-specific vocabulary, as experimented in the work by [35], gave better results, and in the work conducted by Facebook research on RoBERTa [36], increased the performance of the model for domain-specific tasks. Hence, custom biomedical vocabulary was used that follows byte-pair encoding (BPE) [37], [38] learned from PubMed preprocessed text [39], and aligned it with the BERT tokenizer so that it corresponded to the embeddings in the model. For the experimental setup, the maximum sequence length was set to 512, and the same batch size was used for pretraining BERT on general corpora [1]. The learning rate was set to 3e-4, and the model was trained for 450k steps on PubMed and 200k steps on MIMIC-III [40] on NVIDIA A100 GPUs with total warmup steps of 2,000.

Table 2. Statistics of pretraining corpora

| Corpus | Statistics | Domain |
| --- | --- | --- |
| English Wikipedia | >5M articles >3B words | English (General) |
| BooksCorpus | >11,038 books >985M words | General |
| PubMed Abstracts | >22M abstracts >5B words | Biomedical |
| PubMed Central | >3M articles >9.6B words | Biomedical |
| MIMIC-III | >2M health-related notes >0.2B words | Medical |

### 4.2. Finetuning the BERT model

Fine-tuning involves further training the pretrained model on a specific task or domain to enhance its performance for that particular application. The goal was to fine-tune the BERT model on six benchmark tasks available in BLURB [20] and BioBERT [4] repository. We used four preprocessed datasets, i.e., BC4CHEMD, LINNAEUS, and EU-ADR, available in the BioBERT repository. The biomedical and clinical datasets were utilized for unsupervised fine-tuning, adapting the model's representations to perform effectively on various biomedical tasks.

We used two NVIDIA A100 GPUs with parallel training to fine-tune our BERT model on each task. FP16 mixed-precision was also used to speed up training and reduce memory usage. It took approximately 15-20 minutes to fine-tune the model for sequence labeling and sequence classification tasks. For sequence labeling, batch sizes of 16 and 32 were used, and the learning rate was 5e-5. For the JNLPBA [28] corpus, we set a learning rate of 1e-5. For datasets with higher number of examples, a train/validation/test split of about 80/10/10 or 75/15/10 is made, whereas for smaller datasets, the split is about 60/20/20. The named entity recognition datasets were trained between 5 and 25 epochs. On the other hand, for sequence classification, a batch size of 32 and learning rates of 2e-5 and 3e-5 were used. The model was trained for 10 epochs on the ChemProt, DDI-2013 and GAD datasets.

## 5. RESULTS AND ANALYSIS

The results of MedicalBERT were compared with three specialized BERT models presented in Table 3. These models include SciBERT [8], ClinicalBERT [7], and BioBERT [4]. We also compared the results with the original BERT model (column 3) trained on general English corpora and RoBERTa [9] (column 4) to evaluate the performance of the model on NLP tasks and its understanding in biomedical, clinical and scientific terms. The results of MedicalBERT under the various evaluation metrics on sequence labeling, relation extraction and document classification are shown in Table 4. On sequence labeling tasks, BioBERT outperformed the other models on three datasets in the biomedical domain. The performance of the RoBERTa model is comparable to that of the proposed model on LINNAEUS [30], with an F1 score of 87.8. MedicalBERT achieved the best results on five out of eight datasets in the named entity recognition task. Overall, the model outperforms RoBERTa by 2.04 and competitively outperforms BioBERT by 0.75 (mean F1 increase) and SciBERT by 1.4.

The variation in the F1 evaluation metric of the three models and MedicalBERT across the four sequence labeling tasks is depicted in Figure 2(a). Comparing with BioBERT, MedicalBERT achieved better results in relation extraction and sequence classification tasks with similar scores on two datasets (85.6 and 85.5 for GAD and 76.8 and 76.1 for DDI-2013). The model's performance is similarly compared to others and depicted in Figure 2(b). MedicalBERT faced competition with SciBERT on EU-ADR dataset, with micro average F1 scores of 85.6 and 81.2 respectively. The performance of our proposed model on sentence similarity, question answering and document classification is depicted in Figure 2(c). The better results of MedicalBERT compared to the other models for many benchmark datasets reported can be attributed to larger and relevant pre-training data, as compared to ClinicalBERT (which is trained on mainly clinical documents), SciBERT (which is trained on scientific documents and less biomedical text) and RoBERTa (which is trained on general purpose English text), and domain-specific biomedical vocabulary used compared to the other models. The tradeoff caused due to computational cost to train the model on a larger biomedical and clinical corpus found the need to reduce the number of training steps compared to other models in the pre-training stage, and differences in hyperparameters set during the fine-tuning stage with the others make them possible reasons for poorer results in some tasks. Also, the unavailability of proper and high standard datasets for fine-tuning for sequence labeling task, relation extraction and sentence similarity tasks also resulted in a decrease in performance.

Table 3. Performance of BERT and other models on 17 benchmark datasets

| Dataset | Metric | BERT | RoBERTa | SciBERT | ClinicalBERT | BioBERT | MedicalBERT (ours) |
|---|---|---|---|---|---|---|---|
| NCBI-Disease | F1 | 85.6 | 86.6 | 88.2 | 86.3 | **89.1** | 88.3* |
| BC2GM | F1 | 81.8 | 80.9 | 83.4* | 81.7 | **83.8** | 82.8 |
| BC5CDR-Chem | F1 | 91.2 | 90.8 | 92.5 | 90.8 | 92.8* | **93.2** |
| BC5CDR-Disease | F1 | 82.4 | 82.3 | 84.5 | 83.0 | 84.7* | **86.8** |
| JNLPBA | F1 | 90.0 | 90.6 | 91.8* | 90.3 | **92.2** | 91.6 |
| LINNAEUS | F1 | 84.3 | <u>87.8</u> | 84.1 | 84.8 | 86.2* | **87.8** |
| BC4CHEMD | F1 | 74.9 | 80.1 | 79.6 | 78.6 | 80.4* | **83.6** |
| Species-800 | Macro-F1 | 72.3 | 73.0 | 73.1 | 72.1 | 73.2* | **74.3** |
| EBM-PICO | Micro-F1 | 77.7 | 77.7 | 80.9* | 78.4 | 80.9* | **85.8** |
| GAD | F1 | 84.6 | 85.0 | 85.5* | 85.1 | 85.0 | **85.6** |
| EU-ADR | Micro-F1 | 80.0 | 79.5 | 81.2* | 78.2 | 80.9 | **85.4** |
| DDI-2013 | Micro-F1 | 71.9 | 72.9 | 75.2 | 72.0 | 76.1* | **76.8** |
| ChemProt | Accuracy | 82.7 | 81.2 | 87.1 | 91.2* | 89.5 | **92.0** |
| PubMedQA | Accuracy | 51.6 | 52.8 | 57.4 | 49.1 | 60.2* | **61.3** |
| BioASQ | Pearson | 74.4 | 75.2 | 78.9 | 68.5 | 84.1* | **87.9** |
| BIOSSES | Micro-F1 | 80.2 | 79.6 | 81.1 | 80.7 | 81.5* | **81.6** |
| HoC | F1 | 85.6 | 86.6 | 88.2 | 86.3 | **89.1** | 88.3* |
| Total (Mean) | | 79.5 | 80.2 | 81.9 | 79.8 | 82.6* | **84.3** |

Bold scores denote the best, asterisk (*) marks the second-best, and underlined scores indicate ties

Table 4. MedicalBERT results on named entity recognition, relation extraction, and document classification evaluated by F1, recall, and precision

| Dataset | F1 | R (Recall) | P (Precision) |
|---|---|---|---|
| NCBI–Disease | 88.301 | **90.417** | 86.282 |
| BC2GM | 82.754 | **83.304** | 82.212 |
| BC5CDR–Chem | **93.162** | 93.110 | 92.214 |
| BC5CDR-Disease | **86.784** | 86.618 | 84.966 |
| LINNAEUS | **87.822** | 84.042 | 85.365 |
| BC4CHEMD | 83.566 | **84.355** | 82.791 |
| JNLPBA | 91.651 | 89.983 | **92.328** |
| Species–800 | **74.315** | 72.565 | 70.895 |
| DDI–2013 | 76.889 | 75.485 | **78.367** |
| EU–ADR | 85.433 | **86.150** | 74.206 |
| ChemProt | **92.053** | 90.053 | 88.622 |
| GAD | 86.604 | **92.833** | 79.765 |
| HoC | **88.378** | 81.311 | 84.678 |

Bold indicates best scores

MedicalBERT's practical applications in healthcare and biomedical fields are transformative, especially as the model adapts to the complex nuances of medical language. By fine-tuning BERT to recognize and interpret biomedical text, MedicalBERT effectively addresses several critical needs across various domains in healthcare, enhancing decision support, diagnostic accuracy, and research advancements. With the capability to capture context from biomedical text, MedicalBERT facilitates efficient extraction of critical information from large volumes of medical documents. This includes identifying and categorizing symptoms, medications, treatment plans, and lab results from free-text EHRs. Medical information extraction not only assists in patient care but also accelerates research activities by structuring data for use in large-scale studies and meta-analyses. MedicalBERT can contribute to patient safety by identifying potential drug-drug interactions. By training on large datasets of drug interactions and EHRs, MedicalBERT can analyze prescriptions and alert clinicians to dangerous combinations, thus reducing the risk of adverse events in polypharmacy scenarios. Future work may explore further customization and evaluation of MedicalBERT across these applications to optimize its integration into real-world healthcare systems and clinical workflows.

The use of MedicalBERT and other transformer-based models in the biomedical field opens up vast possibilities for improving healthcare and clinical research. However, the sensitive nature of medical data introduces complex ethical and privacy concerns. Biomedical data is inherently sensitive, containing personally identifiable information (PII) and protected health information (PHI) that, if exposed, could have severe consequences for individuals. Given MedicalBERT's reliance on large datasets for effective pre-training and fine-tuning, ensuring data privacy during model development is essential. This can be done by establishing clear guidelines for data anonymization, encryption, and differential privacy techniques in medical AI applications. The use of medical records for training models like MedicalBERT should be accompanied by transparent consent practices. Patients should be informed about how their data may contribute to training models for various applications, along with the potential benefits and risks. Policies requiring regular ethical and fairness audits can ensure that models like MedicalBERT meet high standards of transparency and accountability. As MedicalBERT becomes integrated into clinical workflows, the need for explainability grows. Policy frameworks that mandate explainability in AI-driven decision-making processes can promote accountability and increase model acceptance among clinicians, improving overall trust in the technology.

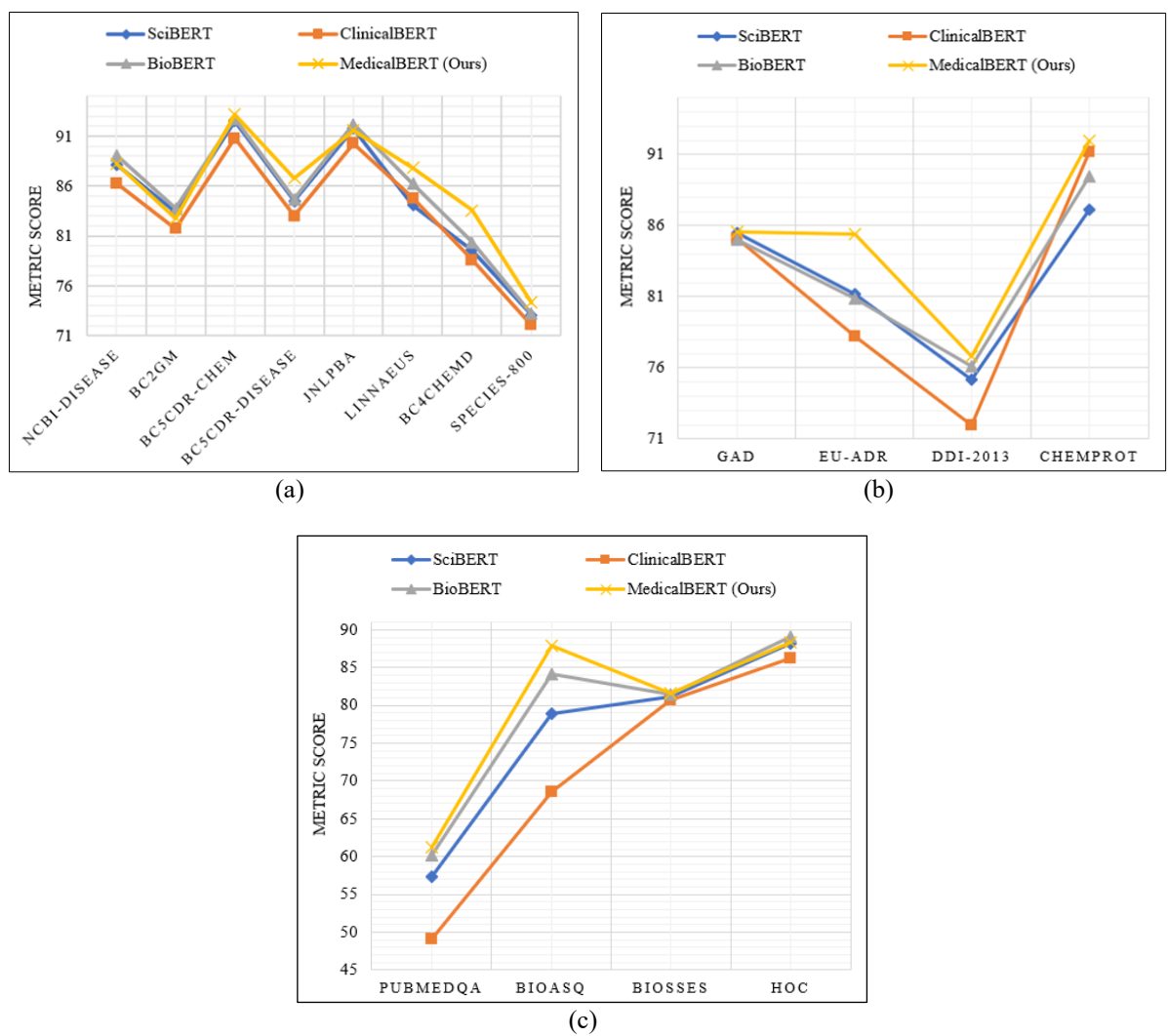

Figure 2. Performance variation of biomedical models and proposed model on (a) sequence labeling; (b) relation extraction; and (c) sentence similarity, question answering and document classification tasks

## 6. CONCLUSION

The BERT model has been subjected to six various benchmark tasks on 17 datasets. These tasks were addressed by using transfer learning techniques. This model outperforms the other specialized BERT models on sequence classification, sentence similarity tasks, question answering and document classification tasks. We plan to train the BERT-based variant on larger biomedical corpora and clinical reports using better computational resources, to determine the effect of training corpora and their length on performance in the future. Additionally, we also plan to evaluate MedicalBERT on the five biomedical text-mining tasks with ten corpora from the biomedical language understanding evaluation (BLUE) benchmark. Furthermore, the future work also includes linguistic support by incorporating training and evaluation on various datasets and texts presented in languages other than English to study the impact on the performance of the model.

## ACKNOWLEDGEMENTS

We acknowledge the support of R. V. College of Engineering for providing the computational resources necessary to complete this research.

FUNDING INFORMATION

Authors state no funding involved.

AUTHOR CONTRIBUTIONS STATEMENT

This journal uses the Contributor Roles Taxonomy (CRediT) to recognize individual author contributions, reduce authorship disputes, and facilitate collaboration.

| Name of Author     | C | M | So | Va | Fo | I | R | D | O | E | Vi | Su | P | Fu |
|--------------------|---|---|----|----|----|---|---|---|---|---|----|----|---|----|
| K. Sahit Reddy     | ✓ | ✓ | ✓  | ✓  | ✓  |   | ✓ |   | ✓ | ✓ | ✓  | ✓  |   |    |
| N. Ragavenderan    | ✓ | ✓ | ✓  | ✓  | ✓  | ✓ |   |   | ✓ | ✓ | ✓  | ✓  |   |    |
| Vasanth K.         | ✓ | ✓ | ✓  |    | ✓  | ✓ |   | ✓ | ✓ | ✓ | ✓  | ✓  |   |    |
| Ganesh N. Naik     | ✓ | ✓ | ✓  |    | ✓  |   | ✓ | ✓ | ✓ | ✓ | ✓  | ✓  |   |    |
| Vishalakshi Prabhu | ✓ |   |    | ✓  |    | ✓ | ✓ |   |   | ✓ |    |    | ✓ | ✓  |
| Nagaraja G. S.     |   | ✓ |    | ✓  |    | ✓ | ✓ |   |   | ✓ |    |    | ✓ | ✓  |

C  : **C**onceptualization        I  : **I**nvestigation           Vi : **Vi**sualization
M  : **M**ethodology              R  : **R**esources               Su : **Su**pervision
So : **So**ftware                 D  : **D**ata Curation           P  : **P**roject administration
Va : **Va**lidation               O  : Writing - **O**riginal Draft   Fu : **Fu**nding acquisition
Fo : **Fo**rmal analysis          E  : Writing - Review & **E**diting

CONFLICT OF INTEREST STATEMENT

Authors state no conflict of interest.

DATA AVAILABILITY

The data that support the findings of this study will be available in https://github.com/ksahitreddy/MedicalBERT following a 2-month embargo from the date of publication to allow for the commercialization of research findings.

## BIOGRAPHIES OF AUTHORS

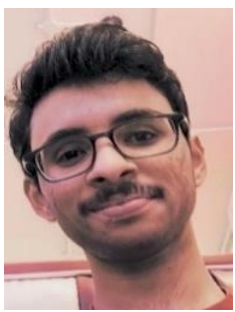

**Mr. K. Sahit Reddy** is a bachelor at R. V. College of Engineering, Bangalore. He holds various certifications in deep learning, cloud computing, and operating systems from Coursera, IBM, Udemy and Infosys Springboard. He has also published a paper as the main author in the 7th IEEE International Conference CSITSS–2023 titled "Traffic data analysis and forecasting" held at R. V. College of Engineering. His technical skills mainly lie in the domain of machine learning, image processing, and cloud computing. He has also taken part as well as volunteered for hackathons held in R. V. College of Engineering. He has also completed his internship training in Women in Cloud Centre of Excellence in R. V. College of Engineering in the year 2023. He is also a member of the Ashwa Mobility Foundation Club, RVCE in the IT subsystem since 2023 and is also a member of ACM–RVCE Student Chapter. He can be contacted at email: sahitreddykangati@gmail.com or ksahitreddy.cs22@rvce.edu.in.

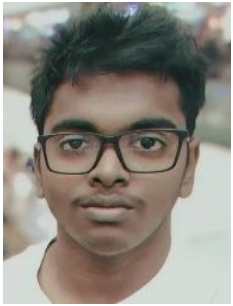

**Mr. N. Ragavenderan** currently is pursuing a B.E. degree at R. V. College of Engineering, Bangalore, is a NTSE Scholar (2020). He has published a paper titled "Traffic data and forecasting" in the 7th IEEE International Conference CSITSS–2023 held at R.V. College of Engineering, Bangalore. Additionally, he serves as the Treasurer of IEEE Special Interest Group on Humanitarian Technology (SIGHT) RVCE and is a member of IEEE Computer Society RVCE and ACM–RVCE Student Chapter. He has organized a National Level Hackathon–"Hack4Soc"–a 24-hour All India Hackathon for Social Impact under IEEE Computer Society, RVCE. He holds certifications from HyperSkill, Udemy, Coursera, and Infosys Springboard. He can be contacted at email: ragava22005@gmail.com or nragavenderan.cs22@rvce.edu.in.

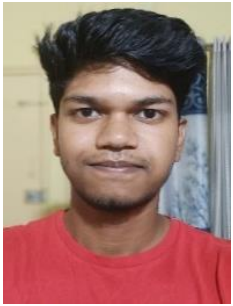

**Mr. Vasanth K.** currently is pursuing a B.E. degree at R. V. College of Engineering, Bangalore, is a KCET (2022) Rank Holder. He holds certification in machine learning from IITKG, NPTEL. He also has other certifications related to Linux, cyber security and networking from CISCO, IBM, Udemy and Infosys Springboard. He has published a paper titled "Traffic data and forecasting" in the 7th IEEE International Conference CSITSS –2023 held at R.V. College of Engineering, Bangalore. He is also part of the Coding Club, RVCE since 2023. He has completed his internship training related to Cloud from Women in Cloud in the year 2023. He can be contacted at email: vasanthkarthikblr@gmail.com or vasanthk.cs22@rvce.edu.in.

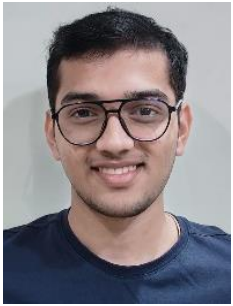

**Mr. Ganesh N. Naik** is a Bachelor of Engineering student at R.V. College of Engineering, Bangalore, presented a paper titled "Traffic data and forecasting" at the prestigious 7th IEEE International Conference CSITSS-2023 and published a paper titled "Classification of underwater mines with convolutional neural networks" at the International Journal of Applied Engineering and Technology (London). He holds the role of Treasurer at the IEEE Computer Society RVCE, a member of IEEE SIGHT and organized "Hack4Soc 2.0", a National Level Hackathon under IEEE Computer Society, RVCE. Mr. Naik has obtained certifications from Udemy, Coursera, and Infosys Springboard, showcasing his commitment to continuous learning. He can be contacted at email: ganeshnaik2106@gmail.com or ganeshnnaik.cs22@rvce.edu.in.

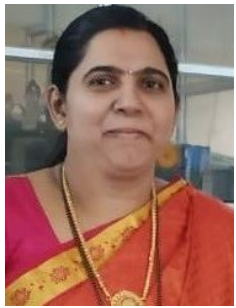

**Dr. Vishalakshi Prabhu** is an Assistant Professor at Rashtreeya Vidyalaya College of Engineering (RVCE), brings extensive expertise with an M.Tech. (Gold Medalist) and Ph.D. degree. With 16 years in teaching and 1 year in industry, she specializes in wireless communications, network security, cloud computing, and algorithms. She has supervised numerous projects, authored over 30 publications in international journals and conferences, and actively collaborates with industry partners. Recognized for her academic excellence, she holds certifications from Coursera, IBM, and UiPath. She can be contacted at email: vishalaprabhu@rvce.edu.in.

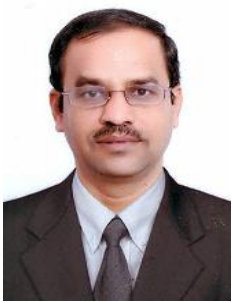

**Dr. Nagaraja G. S.** holds a Ph.D. in computer science and engineering and serves as a Professor and Associate Dean at Rashtreeya Vidyalaya College of Engineering (RVCE) since December 2005. With over 30 years of teaching and 19+ years of research experience, he specializes in computer networks and management, multimedia communications, and protocol design. He has supervised many projects, published extensively, and actively contributes to professional organizations like IEEE and ISTE. His leadership roles, consultancy activities, and dedication to academic excellence underscore his significant contributions to the field. He can be contacted at email: nagarajags@rvce.edu.in.